\documentclass[runningheads]{llncs}
\usepackage[T1]{fontenc}
\usepackage{graphicx}
\usepackage{amsmath}
\usepackage{amssymb}
\usepackage{url}
\usepackage{arydshln}
\usepackage{booktabs}
\usepackage{array}
\usepackage{multirow}

\begin{document}
\title{BRACHIO: Broad Representations Across Centres for Head-and-neck Imaging Oncology}
%
\author{Roxanne De Maeyer\inst{1,2,3}\orcidID{0009-0001-7438-1235}\and Ine Dirks\inst{2,3}\orcidID{0000-0002-1648-0358} \and Yannick Brackenier\inst{1}\orcidID{0000-0002-9210-5939} \and Simon DeKeyser\inst{1}\orcidID{0000-0002-1322-7621} \and Jef Vandemeulebroucke\inst{2,3}\orcidID{0000-0001-5714-3254} \and Vicky De Ridder\inst{1}\orcidID{0009-0003-9080-0676}} 
\authorrunning{R. De Maeyer et al.}
\titlerunning{HECKTOR 2026 Submission of Team nucli}
\institute{Nuclivision bv, Edward Pynaertkaai 87, 9000 Gent, Belgium  \and
Vrije Universiteit Brussel (VUB), Pleinlaan 2, 1050 Brussel, Belgium
\and
imec, Kapeldreef 75, 3001 Leuven, Belgium}
\maketitle              
\begin{abstract}
The HECKTOR 2026 challenge benchmarks machine learning for head and neck cancer, where
teams delineate primary and nodal tumour volumes, perform TN staging, and predict
recurrence-free survival (RFS). This report describes the submissions for the HECKTOR
challenge of team nucli.

As the three tasks span both dense and global information from multicentric PET/CT scans,
we utilise self-supervised pre-trained encoders, training a 2D DINOv2 encoder
(PET-DINOv2-MIPs) on HECKTOR 2026's PET scans. Our final test submission ensembles a
DynUNet trained from scratch with a SwinUNETR whose Swin trunk was initialised from a
SimMIM-pre-trained checkpoint for segmentation, reaching a mean Dice of $0.5409$. For TN staging,
$\text{GTV}_\text{p}$/$\text{GTV}_\text{n}$ geometric features from the segmentation masks
are fit to stage probabilities by logistic regression (LR) and fused with PET-DINOv2-MIPs
deep features (PCA + LR) via a second-level LR, reaching balanced accuracies of $0.3962$
(T-stage) and $0.6432$ (N-stage). For the RFS prognosis task, we fine-tune the last two
blocks of the PET-DINOv2-MIPs ViT-S encoder jointly with an MLP head under a CoxPH loss,
concatenating clinical covariates with the imaging features, reaching a test Harrell's
C-index of $0.6577$ and a weighted challenge score of $0.5802$. The prognosis result is
comparable to the best entry of the previous edition and consistent with our internal
cross-validation, while our segmentation result is well below it and limits the staging
results that depend on it.

\end{abstract}
\section{Introduction}

Treatment options in head and neck (H\&N) cancer considerably impact patient quality of
life, even with positive survival outcomes. Selection of a treatment regime that is
curative yet minimises impact on later quality of life is therefore
essential~\cite{duprezDeescalationHeadandneckCancer2026}, which requires an estimate of the
patient's recurrence risk. Estimating this risk consistently is difficult: current risk
profiling combines noisy factors (self-reported smoking and alcohol history, performance
status, comorbidity, treatment history, and scanner-dependent PET
markers~\cite{boellaardStandardsPETImage2009}) via clinical judgement rather than a fitted
estimator, and not all factors are available in every hospital. Machine learning offers
reproducible risk estimates fitted directly to observed
outcomes~\cite{swansonPatternsPatientsAdvances2023}, but whether these transfer beyond
their development cohort is a separate question that single-centre studies cannot answer.

The HECKTOR 2026 challenge addresses this directly, providing $\approx$700 training cases
from 8 sites, $\approx$50 validation patients from unseen centres, and $\approx$400 test
subjects from three centres (only one overlapping with training), spanning scanner
generations and protocols. It consists of an automated tumour segmentation task and two
image-level tasks, TN staging and recurrence-free survival (RFS) prognosis. Segmentation
provides a label for every voxel of every case. Prognosis provides one time-to-event label
per patient, of which only 148 of 727 in the 2026 training cohort are uncensored events.
From this limited data volume a model must learn both a prognostic representation and
robustness to scanners it has not seen before. The 2025 edition shows the consequence: one
RFS entry reached cross-validation C-index near 0.71, yet fell to 0.53 on the multi-centre
test cohort, while segmentation performance from the same pipeline
held~\cite{saeedHEadNeCKTumOR2026}.

Self-supervised pre-training (SSL) is a natural response, as it consumes no label
information and can exploit the full imaging distribution rather than only the labelled
subset. We therefore decouple representation learning from the downstream objectives.
Because the three tasks differ in what they require of a representation, we pre-train two
encoders: a DINOv2 encoder on 2D maximum-intensity projections (MIPs) of PET, used for the
image-level staging and prognosis tasks, and a Swin transformer masked image model on 3D
PET/CT, to retain the spatial detail for dense segmentation. Our goal is to evaluate how
well these HECKTOR-only representations transfer to the three tasks, and whether they hold
up across centres between internal cross-validation and the held-out test cohort. We do not
claim a controlled measurement of the benefit of pre-training itself; the reported
experiments do not isolate it (Section~\ref{sec:limitations}).

\section{Methods}

\subsection{Data}
All models are trained on HECKTOR 2026 data only, as no external training data was
allowed. The dataset is described in detail in the accompanying data
descriptor~\cite{saeedMultimodalMulticentricHead2025}, and the preceding edition of the
benchmark in~\cite{saeedHEadNeCKTumOR2026}; further task and split details are on the
challenge website: \url{https://hecktor26.grand-challenge.org/}.

\subsection{Self-supervised pre-training methods}
\subsubsection{DINOv2} \label{sec:dino}
is an SSL method known to transfer well to both global
and dense downstream tasks~\cite{baharoonEvaluatingGeneralPurpose2024}. It trains a student
Vision Transformer (ViT) against an EMA teacher on \texttt{CLS} tokens
only~\cite{caronEmergingPropertiesSelfSupervised2021}, giving strong image-level but weaker
dense features. DINOv2~\cite{oquabDINOv2LearningRobust2024} adds an
iBOT~\cite{zhouIBOTImageBERT2022} loss on masked patch tokens, creating features better
suited to dense tasks. We build on DINOv2, as the newer DINOv3~\cite{simeoniDINOv32025} is
designed for a billion-image, billion-parameter regime less suited to the smaller datasets
common in medical imaging. We pre-train on 2D MIPs of the provided 3D PET data, since whole
3D PET volumes tend to be largely sparse. The value of MIPs for global tasks has been shown
in earlier work~\cite{liMultiviewDeepLearning2026a}.

Each PET volume is processed into 4 MIPs from oblique angles ($0^\circ$, $45^\circ$,
$90^\circ$, $135^\circ$), resampled to $2.0$~mm isotropic pixel spacing, rendered at
$480\times480$ and pre-processed with \texttt{log1p}. The backbone is a standard
ViT-S/16~\cite{dosovitskiyImageWorth16x162021} (dim=384, blocks=12, heads=6, head dim=64,
MLP hidden=1536, $\approx 21.5$M parameters), trained on a single NVIDIA A100-80GB for 1600
epochs at batch size $128$, with hyperparameters similar
to~\cite{oquabDINOv2LearningRobust2024}. This encoder is used in Tasks 2 and 3 and is
further referred to as \textit{PET-DINOv2-MIPs}.

\subsubsection{SimMIM}
Masked Autoencoders~\cite{heMaskedAutoencodersAre2021} transfer exceptionally well to dense
tasks such as segmentation, but appear to be outperformed by contrastively pre-trained
encoders on global tasks~\cite{waldOpenMind3DMedical2025}, so we use masked image modelling
for the segmentation task only. We pre-train the Swin transformer encoder of
SwinUNETR~\cite{hatamizadehSwinUNETRSwin2022} ($\texttt{feature\_size}=48$, depths
$(2,2,2,2)$, heads $(3,6,12,24)$) on two-channel (PET, CT) $128^{3}$ volumes under the
SimMIM formulation~\cite{xieSimMIMSimpleFramework2022},
following~\cite{tangSelfSupervisedPreTrainingSwin2022}: $32^{3}$ patches are masked at a
ratio of $0.6$ and reconstructed under an $\ell_2$ loss, with all-background patches
excluded from the loss. Both modalities are resampled to $1.5$~mm isotropic spacing; PET is
transformed with \texttt{log1p}, CT is windowed to $(-200, 200)$~HU and rescaled to
$[0,1]$. Training runs for 800 epochs with Adam (base learning rate $10^{-4}$, effective
batch 8).

\subsection{Task 1: Segmentation}
During the validation phase, leaderboard performance and performance on our in-house
held-out set of 61 patients (random centre-stratified split, seed 42) diverged substantially
across evaluated models, so we adopted an ensembling strategy. Both submitted ensembles
build on a single-stage DynUNet, trained from scratch on the three-class task (background,
GTV$_p$, GTV$_n$) from PET and CT, resampled to 2~mm isotropic spacing, PET clipped to [0,
15]~SUV, CT windowed to $(-200, 200)$~HU, both rescaled to $[0,1]$. Architecture followed
nnU-Net configuration rules~\cite{isenseeNnUNetSelfconfiguringMethod2021} (MONAI 1.4.0
DynUNet): six resolution levels, $3\times3\times3$ kernels, five stride-2 downsamplings,
instance normalisation, residual blocks, deep supervision with three auxiliary decoder
outputs. Training: random $128^3$ patches, 4 per volume, batch size 2 (effective 8), Adam
(lr 0.01, no weight decay, polynomial decay power 0.9), 1000 epochs, augmented with random
affine transforms, axis flips, Gaussian noise, Gaussian smoothing and intensity scaling;
objective softmax Dice over both foreground classes plus voxelwise cross-entropy over all
three. The validation-phase submission paired it with a two-stage variant of identical
configuration (binary lesion mask, 1000 epochs; refined to three classes, 700 epochs, binary
prediction as third input channel).

For the test phase we bundled this DynUNet with one additional model each:
\texttt{test\_sub\_nnU} with a from-scratch residual-encoder nnU-Net, and
\texttt{test\_sub\_Swin} with a SwinUNETR whose Swin trunk was initialised from the SimMIM
checkpoint. Both were trained in a separate framework sharing one data pipeline, five-fold
split and objective; only network and optimizer schedule differed. PET, CT and reference
segmentations were cropped to the superior 400~mm and resampled to 1.5~mm isotropic
resolution; preprocessing matched the DynUNet apart from SUV clipping (20 for nnU-Net, 50
for SwinUNETR). Training used $128^3$ patches, 500 per epoch with one third lesion-centred,
the nnU-Net augmentation set, the same objective, effective batch size two and 500 epochs.
The nnU-Net (six resolution levels, $[32, 64, 96, 128, 128, 128]$ features, $[1, 3, 4, 6, 6,
6]$ residual blocks, deep supervision at every decoder scale, 22.3M parameters) used SGD
(Nesterov momentum 0.99, lr 0.01, weight decay $3 \times 10^{-5}$) under the same polynomial
schedule; the SwinUNETR (62.2M parameters, Swin trunk 8.06M) was fine-tuned end to end with
AdamW (lr $10^{-4}$, weight decay $10^{-5}$), a ten-epoch warm-up and cosine annealing,
without deep supervision.

At inference each member predicts on its own grid with sliding-window inference ($128^3$
windows, 50\% overlap, Gaussian weighting, no test-time augmentation); per-class probability
volumes are resampled onto a shared grid, averaged with equal weight, argmaxed once and
cropped to the superior 400~mm. Due to time constraints, each member is a single-fold model.

\subsection{Task 2: Staging} \label{sec:staging}
TN staging is a multi-class classification problem, evaluated with balanced accuracy, i.e.
recall averaged over the classes of each stage. We combine two classifiers, one on geometric
features from the Task-1 masks and one on PET-DINOv2-MIPs embeddings, the final
classification fused by a second-level logistic regression.

\textbf{\textit{Geometric features (classifier 0a)}} N-stage labels were corrected by the
organising team using AJCC 7th edition criteria~\cite{edgeAmericanJointCommittee2010} on the
ground-truth masks; T-stage labels were not, and remain unreliable: geometric derivation
from ground-truth masks reassigns 151 of 189 provided T1 cases and recovers none of the 118
T4 cases (defined by local invasion rather than $\text{GTV}_\text{p}$ characteristics),
while shifting N-stage counts by at most 16 of 782 patients. We therefore do not derive
stage directly from mask geometry, but extract 12 mask features covering the AJCC size,
count and laterality criteria and fit a logistic regression onto stage probabilities: three
for the primary (presence, greatest diameter of its largest connected component, total
volume), four for node burden (presence, largest node diameter, number of node components,
total volume), and five for criteria a single size threshold misses (second-largest node
diameter, bilaterality flag, number of ipsi- and contralateral nodes, largest contralateral
node diameter). Diameters are maximum Feret diameters, laterality is relative to the
primary's side of the midline, and no minimum-size filtering of nodal components was applied
(threshold of one voxel). On ground-truth masks this gives $0.964$ N-staging balanced
accuracy but only $0.399$ for T.

\textbf{\textit{SSL encoder features (classifier 0b)}} To avoid relying solely on the Task-1
masks, a second, mask-independent classifier uses the frozen PET-DINOv2-MIPs encoder
(Section~\ref{sec:dino}), with PET pre-processed as during pre-training. Each MIP is encoded
separately, yielding $30\times30=900$ patch tokens and one \texttt{CLS} token, averaged into
one $384$-dim token per MIP; the four per-patient tokens are averaged into a final $384$-dim
vector, reduced by PCA to 8 components and mapped to per-class T/N probabilities by a
logistic regression.

\begin{figure}[h!]
    \centering
    \includegraphics[width=0.88\linewidth]{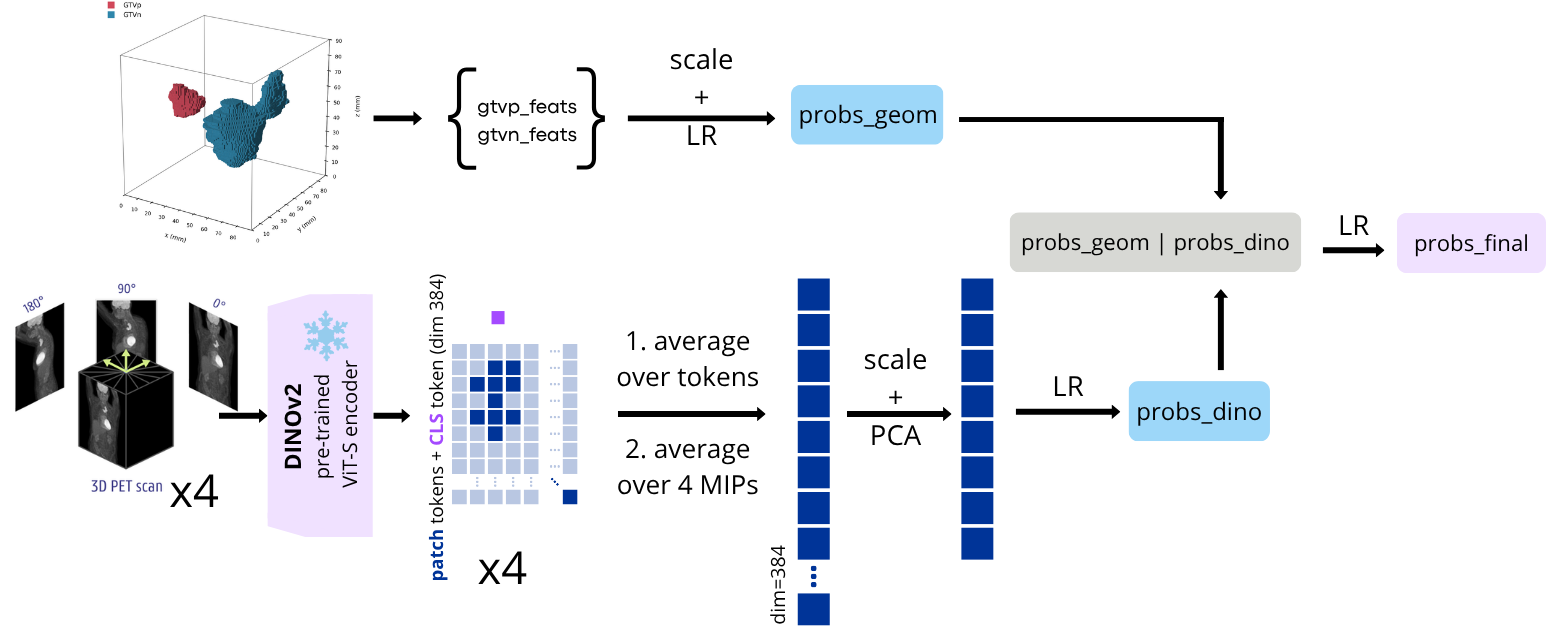}
    \caption{Graphic visualisation of the proposed methodology for Task 2, using a fusion of
    geometric features and extracted PET-DINOv2-MIPs features. LR = logistic regression,
    PCA = principal component analysis feature reduction.}
    \label{fig:staging-pipeline}
\end{figure}

\textbf{\textit{Fusion}} Direct concatenation would let the dense 384-dim imaging block
dominate the regularisation path of the 12 geometric features, so we use \emph{late fusion}
(stacked generalisation~\cite{wolpertStackedGeneralization1992},
Fig.~\ref{fig:staging-pipeline}): each view is fitted independently as a level-0 classifier
and a level-1 logistic regression learns how much to trust each view per class, 5 fold cross-validated. To avoid
leaking in-sample confidence, the level-1 model is fitted exclusively on \emph{out-of-fold}
level-0 probabilities, with standardisation, PCA and the level-0 classifiers refitted per
fold on patients they never saw. For deployment the level-0 models are refitted on the full
training set, while the level-1 model is kept as trained on the out-of-fold matrix. All
three are multinomial logistic regressions (L-BFGS, $\ell_2$ regularisation, class-balanced
weights against the strong imbalance, in particular N2), with $\min(5,\,\min_k n_k)$ folds:
5 for N, 3 for T, the latter limited by the rarity of T0.

\textbf{\textit{Provenance of the segmentation masks}} The out-of-fold protocol applies to
the staging classifiers only, not to the masks they consume: geometric features were
computed from masks inferred for all training patients. Mask quality underlying
Table~\ref{tab:staging-results} is therefore optimistic relative to deployment. The
submitted pipeline uses the mask of its own Task-1 stage, so the geometric features differ
between submissions.

\subsection{Task 3: Prognosis} \label{sec:prognosis}
\begin{figure}[h!]
    \centering
    \includegraphics[width=0.88\linewidth]{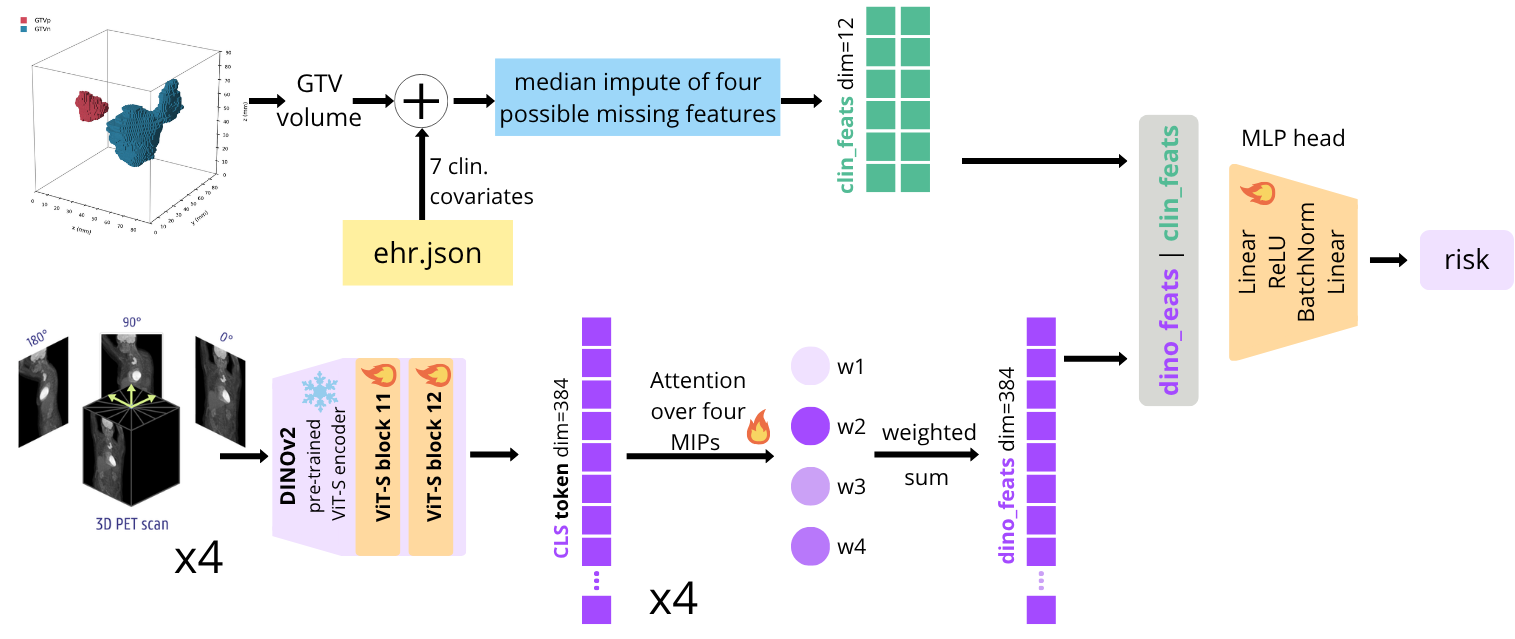}
    \caption{Graphic visualisation of the proposed methodology for Task 3. The submitted
    method was method 2, with a partially fine-tuned DINOv2 encoder, \texttt{CLS}-token
    features and MIP-attention pooling.}
    \label{fig:prognosis}
\end{figure}

The prognosis cohort consisted of $727$ patients with RFS labels, of which $148$ patients
had events / were not censored. Because of the low event count, the batch size was set at
the maximum for compute capacity ($\frac{1}{4}$ of the training fold), stratified by centre
and event.

\textbf{\textit{SSL encoder features}} The prognosis task uses the PET-DINOv2-MIPs encoder
(Section~\ref{sec:dino}) and attaches a Cox survival head. The submitted model uses only the
\texttt{CLS} token of each of the four MIPs; patch tokens and the GTV masks are not used to
pool image features. \texttt{CLS} tokens gave better generalisation than GTV-masked patch
tokens (Table~\ref{tab:prognosis-deep}) and decouple the prognosis branch from segmentation
quality, which was the weakest component of our pipeline. The Task-1 mask therefore enters
the prognosis model only through the derived GTV$_p$ volume covariate.

\textbf{\textit{Clinical covariates}} Clinical covariates per patient are age, gender,
tobacco consumption, alcohol consumption, HPV status, performance status, treatment status
and GTV$_p$ volume (from the Task-1 segmentation mask). Age and GTV$_p$ volume are
continuous, the remaining six are binary. Age was standardised to zero mean and unit
variance using training-fold statistics. Four covariates are only partially observed
(tobacco consumption, alcohol consumption, HPV status and performance status); these were
median-imputed (training-fold only) with a binary missingness indicator appended per
covariate, giving $8+4=12$ clinical features and a $384+12=396$-dim input to the survival
head.

\textbf{\textit{Architecture of the Cox survival head}} The four per-MIP \texttt{CLS} tokens
are combined by MIP-attention pooling: an attention MLP with $\tanh$ activation computes a
normalised weight per MIP, and the weighted sum gives a single 384-dim patient vector. The
survival head maps the fused 396-dim vector through $\texttt{Linear}(396\rightarrow128)$,
$\texttt{ReLU}$, $\texttt{BatchNorm}$, $\texttt{Dropout(0.3)}$, then a bias-free
$\texttt{Linear}(128\rightarrow1)$ outputting a scalar risk (higher risk = shorter RFS). The
Cox PH loss sorts patients by descending time. We tested two paradigms; the complete flow of
the second (submitted) methodology is shown in Fig.~\ref{fig:prognosis}.
\begin{enumerate}
    \item \textit{Frozen encoder.} Features pooled by masked mean over in-GTV patch tokens,
    standardised and reduced by PCA to 16 components, concatenated with the clinical
    covariates (PCA applied to the imaging block only), and fitted with an $\ell_2$-penalised
    CoxPH model~\cite{coxRegressionModelsLifeTables1972} ($\lambda=0.1$, penalised
    partial-likelihood maximisation). Standardisation, PCA and the Cox model are fitted on
    the training fold only and applied unchanged to the validation fold.
    \item \textit{Fine-tuned encoder tail.} The last two blocks and final norm of the
    ViT-S/16 DINOv2 encoder are fine-tuned jointly with the MIP-attention pooler, fusion MLP
    and a bias-free linear Cox head for 140 epochs. Trainable parameters: 3{,}650{,}562
    total, of which 3{,}549{,}696 come from unfreezing the last encoder blocks; the remaining
    17{,}921{,}280 encoder parameters stay frozen.
\end{enumerate}

Both paradigms are evaluated under the same five-fold cross-validation, split randomly and
stratified by event, so all rows of Table~\ref{tab:prognosis-deep} are directly comparable.
No prognosis configuration was cross-validated with centre-grouped folds, so these internal
estimates do not measure cross-centre generalisation
(Section~\ref{sec:limitations}).

\section{Results \& Discussion}

Table~\ref{tab:test-results} reports both submissions on the held-out test cohort. They
differ in the second segmentation ensemble member and in the masks feeding the geometric
staging classifier. \texttt{test\_sub\_Swin} was our final submission and scored higher on
the weighted metric.

\begin{table}[tb]
\centering
\caption{Test-phase leaderboard results. \emph{Weighted} is the challenge's combined score
over the three tasks. Both submissions use the same prognosis model and therefore have the
same C-index.}
\label{tab:test-results}
\small
\setlength{\tabcolsep}{5pt}
\begin{tabular}{@{}l ccccc@{}}
\toprule
\textbf{Submission} & Weighted & Mean Dice & RFS C-index & T bal.\ acc. & N bal.\ acc. \\
\midrule
\texttt{test\_sub\_Swin} (final) & $\mathbf{0.5802}$ & $\mathbf{0.5409}$ & $0.6577$ & $0.3962$ & $\mathbf{0.6432}$ \\
\texttt{test\_sub\_nnU}          & $0.5447$ & $0.4677$ & $0.6577$ & $\mathbf{0.4214}$ & $0.5195$ \\
\bottomrule
\end{tabular}
\end{table}

\subsection{Task 1: Segmentation}
Both test ensembles perform similarly on the 61-patient in-house set. For
\texttt{test\_sub\_nnU} the GTV$_p$ Dice is $0.722$ (false-positive volume $1.37$~ml,
false-negative volume $3.54$~ml) and the GTV$_n$ Dice is $0.684$ ($3.22$~ml, $3.94$~ml); for
\texttt{test\_sub\_Swin} these are $0.736$ ($1.65$~ml, $3.19$~ml) and $0.661$ ($3.91$~ml,
$4.03$~ml). The validation-phase ensemble reached $0.724$ and $0.672$ on the same set. This
in-house ordering does not predict the test ordering, which is expected as the set shares
centres with the training data and was used during model selection.

The mean Dice of $0.5409$ on the test cohort is well below the $0.75$ of the best 2025
entries, and the $0.073$ difference between the two test ensembles is larger than the
in-house differences above. The SimMIM-initialised SwinUNETR ensemble generalised better
than the from-scratch nnU-Net ensemble, but the two differ in architecture and optimizer
schedule as well as in initialisation, so this is not evidence for pre-training. Individual
ensemble members are not reported: they were trained by different authors in two separate
frameworks and were not all evaluated on a common held-out split, so their Dice values are
not comparable.

\subsection{Task 2: Staging}
Table~\ref{tab:staging-results} shows in-house staging results via 3- and 5-fold
cross-validation for T- and N-staging respectively, per segmentation source and per method.
Adding DINO features improved T-staging and slightly degraded N-staging, both in-house and
on the validation leaderboard: with masks from the validation-phase ensemble, fusion reached
$0.418$ (T) and $0.679$ (N) against $0.385$ and $0.686$ for geometric features alone. We
therefore continued with late fusion for the test phase, assuming its added value lies
mainly in generalisability towards unseen centres and in the harder T-staging task.

\begin{table}[h!]
\centering
\caption{Balanced accuracy for T-stage (3-fold) and N-stage (5-fold), comparing geometric
features only (0a), DINO features only (0b) and fusion (level-1). The dino row is identical
across sources by construction, as classifier 0b is mask-independent. Results are on
held-out folds of the training dataset; the masks feeding the geometric features were not
generated out-of-fold (Section~\ref{sec:staging}). Best value per segmentation source and
task in bold.}
\label{tab:staging-results}
\small
\setlength{\tabcolsep}{4pt}
\begin{tabular}{@{}l cc @{\hspace{1.2em}} cc @{\hspace{1.2em}} cc@{}}
\toprule
 & \multicolumn{2}{c}{\texttt{ground-truth}} & \multicolumn{2}{c}{\texttt{test\_sub\_nnU}} & \multicolumn{2}{c}{\texttt{test\_sub\_Swin}} \\
\cmidrule(lr){2-3}\cmidrule(lr){4-5}\cmidrule(lr){6-7}
\textbf{Model} & T & N & T & N & T & N \\
\midrule
dino   & 0.251 & 0.278 & 0.251 & 0.278 & 0.251 & 0.278 \\
geom   & \textbf{0.399} & \textbf{0.964} & \textbf{0.410} & \textbf{0.676} & \textbf{0.417} & \textbf{0.694} \\
fusion & 0.366 & 0.963 & 0.395 & 0.663 & 0.403 & 0.675 \\
\bottomrule
\end{tabular}
\end{table}

On the test cohort, T-staging reached $0.3962$ (\texttt{test\_sub\_Swin}) and $0.4214$
(\texttt{test\_sub\_nnU}), close to the $\approx 0.40$ that the geometric classifier reaches
on ground-truth masks. T-staging is therefore limited by the label definition rather than by
segmentation quality: T4 is defined by local invasion and is not recoverable from a mask,
and the provided T labels were not corrected against the masks. N-staging follows mask
quality more closely: $0.6432$ for the ensemble with the higher test Dice against $0.5195$
for the other, a difference of $0.124$ balanced accuracy for a $0.073$ difference in mean
Dice. Compared to the $0.964$ reached on ground-truth masks, N-staging is limited almost
entirely by Task-1 performance rather than by the classifier.

\subsection{Task 3: Prognosis}
Table~\ref{tab:prognosis-deep} shows the internal cross-validation results of the deep Cox
head. Method 1 (frozen DINOv2 features, PCA, linear Cox model) reached maximum Harrell's
C-indices of $0.609\pm0.008$ (clinical only), $0.628\pm0.026$ (DINOv2 only) and
$0.629\pm0.022$ (clinical + DINOv2) at 16 PCA components. Replacing the linear model by a
deep Cox head on the same frozen features raised this to $0.688\pm0.042$, indicating that
the prognostic signal in the DINOv2 features is not well captured by a linear readout.
he prognostic signal in the DINOv2 features is not well captured by a linear readout.
\begin{table}[t]
\centering
\caption{Recurrence-free survival, internal cross-validation. All variants use the same
PET-DINOv2-MIPs encoder followed by MIP-attention pooling, clinical fusion and a Cox head,
and are evaluated under the same event-stratified five-fold cross-validation with random
splits, so all rows are directly comparable. Values are mean $\pm$ standard deviation over
five folds. The prognostic signal in the DINOv2 features is not well captured by a linear
readout.}
\label{tab:prognosis-deep}
\small
\begin{tabular}{@{}l @{\hspace{1.5em}} c@{}}
\toprule
\textbf{Model} & \textbf{Harrell's C $\uparrow$} \\
\midrule
Clinical-only CoxPH (reference) & $0.609 \pm 0.008$ \\
\addlinespace
\multicolumn{2}{@{}l}{\textit{Frozen encoder, deep Cox probe}} \\
\quad GTV latent cross-attention, MLP head & $\mathbf{0.688 \pm 0.042}$ \\
\quad GTV masked mean pool, MLP head & $0.675 \pm 0.044$ \\
\addlinespace
\multicolumn{2}{@{}l}{\textit{Fine-tuned encoder tail (last 2 blocks + norm)}} \\
\quad \texttt{CLS} tokens, MIP attention \textbf{(submitted)} & $0.646 \pm 0.028$ \\
\quad GTV latent cross-attention, MLP head & $0.553 \pm 0.100$ \\
\midrule
\textbf{Validation leaderboard} (submitted) & $0.8893$\textsuperscript{$\S$} \\
\textbf{Test leaderboard} (submitted) & $0.6577$ \\
\bottomrule
\end{tabular}

\vspace{2pt}
{\footnotesize
\textsuperscript{$\S$}\,\textit{Computed on 50 validation patients with an undisclosed
number of events.}\par}
\end{table}
Fine-tuning the encoder tail did not improve on the frozen encoder: at $0.646\pm0.028$ it
falls below the best frozen configuration ($0.688\pm0.042$) under the same folds. We
nonetheless submitted the fine-tuned model, as it gave the highest validation-phase
leaderboard score at the time of submission, and the shipped checkpoint was fold 0 at
$0.710$. On the test cohort it reached $0.6577$, close to the cross-validation mean and
within the fold-to-fold spread, while the validation-leaderboard C-index of $0.8893$
overestimated test performance by $0.23$. Internal cross-validation was therefore a better
predictor of test performance than the 50-patient validation leaderboard, which is also the
basis on which the submitted configuration was chosen over the better-scoring frozen probe.
The test C-index is comparable to the best RFS entry of the 2025 edition
($0.66$)~\cite{saeedHEadNeCKTumOR2026}, obtained here from PET MIPs and clinical covariates
only, without dose maps, without CT, and without using the segmentation masks to pool image
features.

\subsection{Limitations} \label{sec:limitations}
Most of our development effort went to the self-supervised pre-training pipeline and the
prognosis task, and comparatively little to the segmentation methodology. The test Dice of
$0.5409$, against $0.75$ for the best 2025 entries, reflects this. Each ensemble member was
submitted as a single-fold checkpoint instead of a 5-fold ensemble, only two members were
combined per submission, and no test-time augmentation was used. Previous editions have been
won by more extensively ensembled pipelines, so ensembling across folds and architectures
would likely recover part of this gap.

Four choices limit the internal evidence. The 61-patient in-house set was split randomly
with centre stratification rather than by centre, and was used during ensemble selection, so
it is a development set. The masks used to fit classifier 0a were inferred for all training
patients, including patients the segmentation models had seen during training. The submitted
prognosis checkpoint is fold 0, selected as the best of five folds; retraining on all
patients or ensembling the folds would have been a possible improvement. Finally, no
prognosis configuration was cross-validated with centre-grouped folds, so none of our
internal estimates measures cross-centre generalisation, despite that being a stated
motivation of this work; that the test C-index landed close to the cross-validation mean is
therefore reassuring but not something the internal protocol was designed to predict.

The experiments also do not isolate the contribution of pre-training itself, which would
require the same architecture, schedule and CV regime with and without SSL initialisation.
The SimMIM-initialised SwinUNETR differs from the from-scratch models in architecture and
optimizer, the DINOv2 staging features are compared against geometric features rather than
against a randomly initialised encoder, and the prognosis comparisons vary the head as well.
Finally, the provided T-stage labels were not cleaned, and CT was not used for staging or
prognosis despite being available.

\section{Conclusion}
We evaluated self-supervised pre-training for three head and neck oncology tasks in HECKTOR
2026. A PET-DINOv2-MIPs encoder, pre-trained exclusively on HECKTOR data, contributed
features to both TN staging and RFS prognosis, while a SimMIM-pre-trained Swin transformer
supported segmentation. Our final submission reached a weighted score of $0.5802$, with a
mean Dice of $0.5409$, balanced accuracies of $0.3962$ (T) and $0.6432$ (N), and an RFS
C-index of $0.6577$. The prognosis result lies within the spread of our internal
cross-validation and is comparable to the best entry of the previous edition, obtained from
PET MIPs and clinical covariates alone. Segmentation was below the state of the art and
limited N-staging, while T-staging is bounded by the label definition rather than by our
models.

Future work includes ensembling segmentation predictions across all folds and architectures,
incorporating CT into the staging and prognosis pipelines alongside PET, a matched
pre-trained versus randomly initialised comparison to isolate the contribution of
pre-training, centre-grouped cross-validation for the prognosis branch, and extending
self-supervised pre-training to 3D DINO-style models on PET/CT to better exploit volumetric
context. Overall, the self-supervised pre-trained encoders showed promising transfer
performance on the image-level tasks, with the prognosis branch generalising from internal
cross-validation to the multi-centre test cohort.

\begin{credits}
\subsubsection{\ackname} This study was funded with a Baekeland mandate from VLAIO, grant number HBC.2025.0912.

\subsubsection{\discintname}
Y. Brackenier, V. De Ridder and S. DeKeyser are employed by the company Nuclivision. R. De Maeyer is a PhD candidate co-funded by Nuclivision.

\end{credits}
%
%
\bibliographystyle{splncs04}
\bibliography{HECKTOR26.bib}

@misc{baharoonEvaluatingGeneralPurpose2024,
  title = {Evaluating {{General Purpose Vision Foundation Models}} for {{Medical Image Analysis}}: {{An Experimental Study}} of {{DINOv2}} on {{Radiology Benchmarks}}},
  shorttitle = {Evaluating {{General Purpose Vision Foundation Models}} for {{Medical Image Analysis}}},
  author = {Baharoon, Mohammed and Qureshi, Waseem and Ouyang, Jiahong and Xu, Yanwu and Aljouie, Abdulrhman and Peng, Wei},
  year = 2024,
  month = sep,
  number = {arXiv:2312.02366},
  eprint = {2312.02366},
  primaryclass = {cs.CV},
  publisher = {arXiv},
  doi = {10.48550/arXiv.2312.02366},
  urldate = {2026-06-17},
  archiveprefix = {arXiv}
}

@article{boellaardStandardsPETImage2009,
  title = {Standards for {{PET}} Image Acquisition and Quantitative Data Analysis},
  author = {Boellaard, Ronald},
  year = 2009,
  month = may,
  journal = {Journal of Nuclear Medicine: Official Publication, Society of Nuclear Medicine},
  volume = {50 Suppl 1},
  pages = {11S-20S},
  issn = {0161-5505},
  doi = {10.2967/jnumed.108.057182},
  langid = {english},
  pmid = {19380405}
}

@misc{caronEmergingPropertiesSelfSupervised2021,
  title = {Emerging {{Properties}} in {{Self-Supervised Vision Transformers}}},
  author = {Caron, Mathilde and Touvron, Hugo and Misra, Ishan and J{\'e}gou, Herv{\'e} and Mairal, Julien and Bojanowski, Piotr and Joulin, Armand},
  year = 2021,
  month = may,
  number = {arXiv:2104.14294},
  eprint = {2104.14294},
  primaryclass = {cs},
  publisher = {arXiv},
  doi = {10.48550/arXiv.2104.14294},
  urldate = {2026-04-21},
  archiveprefix = {arXiv}
}

@article{coxRegressionModelsLifeTables1972,
  title = {Regression {{Models}} and {{Life-Tables}}},
  author = {Cox, D. R.},
  year = 1972,
  month = jan,
  journal = {Journal of the Royal Statistical Society Series B: Statistical Methodology},
  volume = {34},
  number = {2},
  pages = {187--202},
  issn = {1369-7412, 1467-9868},
  doi = {10.1111/j.2517-6161.1972.tb00899.x},
  urldate = {2026-06-25},
  copyright = {https://academic.oup.com/journals/pages/open\_access/funder\_policies/chorus/standard\_publication\_model},
  langid = {english}
}

@misc{dosovitskiyImageWorth16x162021,
  title = {An {{Image}} Is {{Worth}} 16x16 {{Words}}: {{Transformers}} for {{Image Recognition}} at {{Scale}}},
  shorttitle = {An {{Image}} Is {{Worth}} 16x16 {{Words}}},
  author = {Dosovitskiy, Alexey and Beyer, Lucas and Kolesnikov, Alexander and Weissenborn, Dirk and Zhai, Xiaohua and Unterthiner, Thomas and Dehghani, Mostafa and Minderer, Matthias and Heigold, Georg and Gelly, Sylvain and Uszkoreit, Jakob and Houlsby, Neil},
  year = 2021,
  month = jun,
  number = {arXiv:2010.11929},
  eprint = {2010.11929},
  primaryclass = {cs},
  publisher = {arXiv},
  doi = {10.48550/arXiv.2010.11929},
  urldate = {2026-04-21},
  archiveprefix = {arXiv}
}

@article{duprezDeescalationHeadandneckCancer2026,
  title = {De-Escalation of Head-and-Neck Cancer Radiotherapy: Balancing Cure and Side Effects},
  shorttitle = {De-Escalation of Head-and-Neck Cancer Radiotherapy},
  author = {Duprez, Fr{\'e}deric and Daisne, Jean-Fran{\c c}ois and Longton, Eleonore and Nuyts, Sandra},
  year = 2026,
  journal = {Frontiers in Oncology},
  volume = {16},
  pages = {1798321},
  issn = {2234-943X},
  doi = {10.3389/fonc.2026.1798321},
  langid = {english},
  pmcid = {PMC13105910},
  pmid = {42038372}
}

@article{edgeAmericanJointCommittee2010,
  title = {The {{American Joint Committee}} on {{Cancer}}: The 7th Edition of the {{AJCC}} Cancer Staging Manual and the Future of {{TNM}}},
  shorttitle = {The {{American Joint Committee}} on {{Cancer}}},
  author = {Edge, Stephen B. and Compton, Carolyn C.},
  year = 2010,
  month = jun,
  journal = {Annals of Surgical Oncology},
  volume = {17},
  number = {6},
  pages = {1471--1474},
  issn = {1534-4681},
  doi = {10.1245/s10434-010-0985-4},
  langid = {english},
  pmid = {20180029}
}

@misc{hatamizadehSwinUNETRSwin2022,
  title = {Swin {{UNETR}}: {{Swin Transformers}} for {{Semantic Segmentation}} of {{Brain Tumors}} in {{MRI Images}}},
  shorttitle = {Swin {{UNETR}}},
  author = {Hatamizadeh, Ali and Nath, Vishwesh and Tang, Yucheng and Yang, Dong and Roth, Holger and Xu, Daguang},
  year = 2022,
  month = jan,
  number = {arXiv:2201.01266},
  eprint = {2201.01266},
  primaryclass = {eess.IV},
  publisher = {arXiv},
  doi = {10.48550/arXiv.2201.01266},
  urldate = {2026-08-07},
  archiveprefix = {arXiv}
}

@misc{heMaskedAutoencodersAre2021,
  title = {Masked {{Autoencoders Are Scalable Vision Learners}}},
  author = {He, Kaiming and Chen, Xinlei and Xie, Saining and Li, Yanghao and Doll{\'a}r, Piotr and Girshick, Ross},
  year = 2021,
  month = dec,
  number = {arXiv:2111.06377},
  eprint = {2111.06377},
  primaryclass = {cs.CV},
  publisher = {arXiv},
  doi = {10.48550/arXiv.2111.06377},
  urldate = {2026-05-20},
  archiveprefix = {arXiv}
}

@article{isenseeNnUNetSelfconfiguringMethod2021,
  title = {{{nnU-Net}}: A Self-Configuring Method for Deep Learning-Based Biomedical Image Segmentation},
  shorttitle = {{{nnU-Net}}},
  author = {Isensee, Fabian and Jaeger, Paul F. and Kohl, Simon A. A. and Petersen, Jens and {Maier-Hein}, Klaus H.},
  year = 2021,
  month = feb,
  journal = {Nature Methods},
  volume = {18},
  number = {2},
  pages = {203--211},
  issn = {1548-7105},
  doi = {10.1038/s41592-020-01008-z},
  langid = {english},
  pmid = {33288961}
}

@article{liMultiviewDeepLearning2026a,
  title = {Multi-View Deep Learning for Automated Lymphoma Staging from {{18F-FDG PET}}/{{CT}}: Physician-Level Accuracy with High-Throughput Workflow},
  shorttitle = {Multi-View Deep Learning for Automated Lymphoma Staging from {{18F-FDG PET}}/{{CT}}},
  author = {Li, Duoduo and Li, Ruikun and Ma, Yutong and Huang, Shengyun and Han, Miaofei and Gao, Yaozong and Liang, Ying},
  year = 2026,
  month = feb,
  journal = {EJNMMI Research},
  volume = {16},
  number = {1},
  pages = {48},
  issn = {2191-219X},
  doi = {10.1186/s13550-025-01357-w},
  urldate = {2026-05-06},
  langid = {english}
}

@misc{oquabDINOv2LearningRobust2024,
  title = {{{DINOv2}}: {{Learning Robust Visual Features}} without {{Supervision}}},
  shorttitle = {{{DINOv2}}},
  author = {Oquab, Maxime and Darcet, Timoth{\'e}e and Moutakanni, Th{\'e}o and Vo, Huy and Szafraniec, Marc and Khalidov, Vasil and Fernandez, Pierre and Haziza, Daniel and Massa, Francisco and {El-Nouby}, Alaaeldin and Assran, Mahmoud and Ballas, Nicolas and Galuba, Wojciech and Howes, Russell and Huang, Po-Yao and Li, Shang-Wen and Misra, Ishan and Rabbat, Michael and Sharma, Vasu and Synnaeve, Gabriel and Xu, Hu and Jegou, Herv{\'e} and Mairal, Julien and Labatut, Patrick and Joulin, Armand and Bojanowski, Piotr},
  year = 2024,
  month = feb,
  number = {arXiv:2304.07193},
  eprint = {2304.07193},
  primaryclass = {cs},
  publisher = {arXiv},
  doi = {10.48550/arXiv.2304.07193},
  urldate = {2026-05-07},
  archiveprefix = {arXiv}
}

@misc{saeedHEadNeCKTumOR2026,
  title = {{{HEad}} and {{neCK TumOR}} ({{HECKTOR}}) 2025: {{Benchmark}} of {{Segmentation}}, {{Diagnosis}}, and {{Prognosis}} in {{Multimodal PET}}/{{CT}}},
  shorttitle = {{{HEad}} and {{neCK TumOR}} ({{HECKTOR}}) 2025},
  author = {Saeed, Numan and Hassan, Salma and Hardan, Shahad and Cai, Lishan and Liang, Xinglong and Mazher, Moona and Qayyum, Abdul and Bu, Yansong and Lyu, Mengye and Lin, Yue and Meng, Mingyuan and Huang, Chuanyi and Wang, Lisheng and Chamseddine, Dalal and Ahrari, Shamimeh and Wu, Beining and Chen, Yifei and Mao, Fuyou and Zhang, Hao and Zhao, Baixiang and Ray, Surajit and Guo, Muzi and Xiang, Lei and Dexl, Jakob and Ingrisch, Michael and Depeursinge, Adrien and Rahmim, Arman and Hatt, Mathieu and Andrearczyk, Vincent and Yaqub, Mohammad},
  year = 2026,
  month = jun,
  number = {arXiv:2606.20143},
  eprint = {2606.20143},
  primaryclass = {cs.CV},
  publisher = {arXiv},
  doi = {10.48550/arXiv.2606.20143},
  urldate = {2026-08-06},
  archiveprefix = {arXiv}
}

@misc{saeedMultimodalMulticentricHead2025,
  title = {A {{Multimodal}} and {{Multi-centric Head}} and {{Neck Cancer Dataset}} for {{Segmentation}}, {{Diagnosis}} and {{Outcome Prediction}}},
  author = {Saeed, Numan and Hassan, Salma and Hardan, Shahad and Aly, Ahmed and Taratynova, Darya and Nawaz, Umair and Khan, Ufaq and Ridzuan, Muhammad and Andrearczyk, Vincent and Depeursinge, Adrien and Xie, Yutong and Eugene, Thomas and Metz, Rapha{\"e}l and Dore, M{\'e}lanie and Delpon, Gregory and Papineni, Vijay Ram Kumar and Wahid, Kareem and Dede, Cem and Ali, Alaa Mohamed Shawky and Sjogreen, Carlos and Naser, Mohamed and Fuller, Clifton D. and Oreiller, Valentin and Jreige, Mario and Prior, John O. and Rest, Catherine Cheze Le and Tankyevych, Olena and Decazes, Pierre and Ruan, Su and {Tanadini-Lang}, Stephanie and Valli{\`e}res, Martin and Elhalawani, Hesham and Abgral, Ronan and Floch, Romain and Kerleguer, Kevin and Schick, Ulrike and Mauguen, Maelle and Bourhis, David and Leclere, Jean-Christophe and Sambourg, Amandine and Rahmim, Arman and Hatt, Mathieu and Yaqub, Mohammad},
  year = 2025,
  month = sep,
  number = {arXiv:2509.00367},
  eprint = {2509.00367},
  primaryclass = {cs.CV},
  publisher = {arXiv},
  doi = {10.48550/arXiv.2509.00367},
  urldate = {2026-08-06},
  archiveprefix = {arXiv}
}

@misc{simeoniDINOv32025,
  title = {{{DINOv3}}},
  author = {Sim{\'e}oni, Oriane and Vo, Huy V. and Seitzer, Maximilian and Baldassarre, Federico and Oquab, Maxime and Jose, Cijo and Khalidov, Vasil and Szafraniec, Marc and Yi, Seungeun and Ramamonjisoa, Micha{\"e}l and Massa, Francisco and Haziza, Daniel and Wehrstedt, Luca and Wang, Jianyuan and Darcet, Timoth{\'e}e and Moutakanni, Th{\'e}o and Sentana, Leonel and Roberts, Claire and Vedaldi, Andrea and Tolan, Jamie and Brandt, John and Couprie, Camille and Mairal, Julien and J{\'e}gou, Herv{\'e} and Labatut, Patrick and Bojanowski, Piotr},
  year = 2025,
  month = aug,
  number = {arXiv:2508.10104},
  eprint = {2508.10104},
  primaryclass = {cs},
  publisher = {arXiv},
  doi = {10.48550/arXiv.2508.10104},
  urldate = {2026-05-12},
  archiveprefix = {arXiv}
}

@article{swansonPatternsPatientsAdvances2023,
  title = {From Patterns to Patients: {{Advances}} in Clinical Machine Learning for Cancer Diagnosis, Prognosis, and Treatment},
  shorttitle = {From Patterns to Patients},
  author = {Swanson, Kyle and Wu, Eric and Zhang, Angela and Alizadeh, Ash A. and Zou, James},
  year = 2023,
  month = apr,
  journal = {Cell},
  volume = {186},
  number = {8},
  pages = {1772--1791},
  issn = {1097-4172},
  doi = {10.1016/j.cell.2023.01.035},
  langid = {english},
  pmid = {36905928}
}

@misc{tangSelfSupervisedPreTrainingSwin2022,
  title = {Self-{{Supervised Pre-Training}} of {{Swin Transformers}} for {{3D Medical Image Analysis}}},
  author = {Tang, Yucheng and Yang, Dong and Li, Wenqi and Roth, Holger and Landman, Bennett and Xu, Daguang and Nath, Vishwesh and Hatamizadeh, Ali},
  year = 2022,
  month = mar,
  number = {arXiv:2111.14791},
  eprint = {2111.14791},
  primaryclass = {cs.CV},
  publisher = {arXiv},
  doi = {10.48550/arXiv.2111.14791},
  urldate = {2026-08-07},
  archiveprefix = {arXiv}
}

@misc{waldOpenMind3DMedical2025,
  title = {An {{OpenMind}} for {{3D}} Medical Vision Self-Supervised Learning},
  author = {Wald, Tassilo and Ulrich, Constantin and Suprijadi, Jonathan and Ziegler, Sebastian and Nohel, Michal and Peretzke, Robin and K{\"o}hler, Gregor and {Maier-Hein}, Klaus H.},
  year = 2025,
  month = apr,
  number = {arXiv:2412.17041},
  eprint = {2412.17041},
  primaryclass = {cs.CV},
  publisher = {arXiv},
  doi = {10.48550/arXiv.2412.17041},
  urldate = {2026-06-18},
  archiveprefix = {arXiv}
}

@article{wolpertStackedGeneralization1992,
  title = {Stacked Generalization},
  author = {Wolpert, David H.},
  year = 1992,
  month = jan,
  journal = {Neural Networks},
  volume = {5},
  number = {2},
  pages = {241--259},
  issn = {0893-6080},
  doi = {10.1016/S0893-6080(05)80023-1},
  urldate = {2026-08-06}
}

@inproceedings{xieSimMIMSimpleFramework2022,
  title = {{{SimMIM}}: A {{Simple Framework}} for {{Masked Image Modeling}}},
  shorttitle = {{{SimMIM}}},
  booktitle = {2022 {{IEEE}}/{{CVF Conference}} on {{Computer Vision}} and {{Pattern Recognition}} ({{CVPR}})},
  author = {Xie, Zhenda and Zhang, Zheng and Cao, Yue and Lin, Yutong and Bao, Jianmin and Yao, Zhuliang and Dai, Qi and Hu, Han},
  year = 2022,
  month = jun,
  pages = {9643--9653},
  issn = {2575-7075},
  doi = {10.1109/CVPR52688.2022.00943},
  urldate = {2026-08-03}
}

@misc{zhouIBOTImageBERT2022,
  title = {{{iBOT}}: {{Image BERT Pre-Training}} with {{Online Tokenizer}}},
  shorttitle = {{{iBOT}}},
  author = {Zhou, Jinghao and Wei, Chen and Wang, Huiyu and Shen, Wei and Xie, Cihang and Yuille, Alan and Kong, Tao},
  year = 2022,
  month = jan,
  number = {arXiv:2111.07832},
  eprint = {2111.07832},
  primaryclass = {cs.CV},
  publisher = {arXiv},
  doi = {10.48550/arXiv.2111.07832},
  urldate = {2026-05-20},
  archiveprefix = {arXiv}
}

\end{document}